\documentclass[final]{cvpr}

\usepackage{times}
\usepackage{epsfig}
\usepackage{graphicx}
\usepackage{amsmath}
\usepackage{amssymb}
\usepackage{standalone}
\usepackage{mathptmx}
\usepackage{times}
\usepackage{epsfig}
\usepackage{graphicx}
\usepackage{amsmath}
\usepackage{amssymb}
\usepackage[]{multirow}
\usepackage[nocompress]{cite}

\usepackage{array}
\usepackage{booktabs}
\usepackage{scalefnt}
\usepackage{enumitem}
\usepackage{scalerel}
\usepackage[font=small]{caption}
\DeclareMathOperator*{\argmin}{argmin} 
\DeclareMathOperator*{\argmax}{argmax}

\usepackage[pagebackref=true,breaklinks=true,colorlinks,bookmarks=false]{hyperref}

\def\cvprPaperID{9} 
\def\confYear{CVPR 2021}
\begin{document}

\title{Super-Resolution Appearance Transfer for 4D Human Performances}

%
%

%
%
\author{Marco Pesavento \hspace{3cm} Marco Volino \hspace{3cm} Adrian Hilton \\ 
Centre for Vision, Speech and Signal Processing\\
University of Surrey, UK\\
{\tt\small \{m.pesavento,m.volino,a.hilton\}@surrey.ac.uk}
}


\maketitle

\begin{abstract}
A common problem in the 4D reconstruction of people from multi-view video is the quality of the captured dynamic texture appearance which depends on both the camera resolution and capture volume. Typically the requirement to frame cameras to capture the volume of a dynamic performance ($>50m^3$)  results in the person occupying only a small proportion $<$ 10\% of the field of view. Even with ultra high-definition 4k video acquisition this results in sampling the person at less-than standard definition 0.5k video resolution resulting in low-quality rendering.
In this paper we propose a solution to this problem through super-resolution appearance transfer from a static high-resolution appearance capture rig using digital stills cameras ($>$ 8k) to capture the person in a small volume ($<8m^3$). 
A pipeline is proposed for super-resolution appearance transfer from high-resolution static capture to dynamic video performance capture to produce super-resolution dynamic textures. This addresses two key problems: colour mapping between different camera systems; and dynamic texture map super-resolution using a learnt model.
Comparative evaluation demonstrates a significant qualitative and quantitative improvement in rendering the 4D performance capture with super-resolution dynamic texture appearance. The proposed approach 
reproduces the high-resolution detail of the static capture whilst maintaining the appearance dynamics of the captured video. 
\end{abstract}
\section{Introduction}
\noindent
The increasing popularity of VR/AR technologies has driven a rise in the demand for 4D modelling techniques that accurately reconstruct human performances. 
Video-based capture systems have been developed recently to replace marker-based motion capture enabling the reconstruction of the detailed surface dynamics for complex non-rigid shapes such as people. 
These systems consist of a set of cameras that capture a scene from multiple viewpoints~\cite{collet2015high, de2008performance, starck2007surface}. A 4D reconstruction of the performance of the captured scene can then be retrieved by applying stereo matching and feature tracking approaches. This reconstruction is a sequence of 3D geometric objects of the performer, each with its own pose and texture appearance. Unless the acquisition system has a large number of cameras~\cite{collet2015high, orts2016holoportation, joo2015panoptic, vlasic2009dynamic, guo2019relightables} or the overall volume to capture is limited, the surface texture resolution resolution is limited~\cite{huang2018deep}. 
In general, the resolution of the captured models decreases proportionally to the increasing of the capture volume size. Moreover, the colours of the appearance are significantly influenced by the lighting of the capture space, which can deteriorate the appearance if the illumination is not thoroughly controlled.
\begin{figure}[t!]
\centering
\resizebox{1.\linewidth}{!}{
\includegraphics[width=1.in]{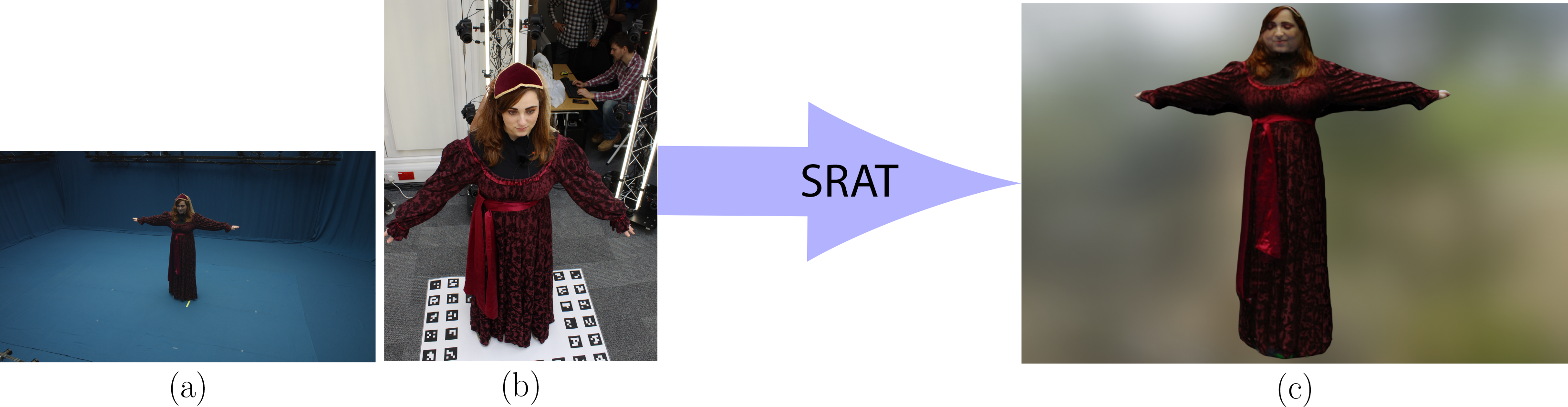}}
\vspace{-2.3em}
\caption{Overview of the SR transfer pipeline: a) Low-resolution capture frame; b) High-resolution image; c) SR result.}
\vspace{-2.2em}
\label{fig:firstim}
\end{figure}
\\Whilst 4D reconstruction systems still face a number of limitations, there are alternative methods that accurately provide a static 3D shape reconstruction from multiple high-resolution DSLR camera images in a limited capture volume giving higher detail reconstruction of surface shape and appearance~\cite{Xie_2019_ICCV,bhatnagar2019multi,lahner2018deepwrinkles,Bi_2020_CVPR}. The resolution of shape detail reconstruction is dependent on the image resolution and the 3D reconstruction systems use high-resolution (HR) cameras that can acquire higher resolution images of the subject. The quality of the reconstruction is highly influenced by the capture volume, which is usually much smaller for 3D reconstruction since the scene to capture is static. The colour response of the acquisition cameras depends on the capture environment, which is designed considering the objective of the capture. The system settings of the two reconstructions differ due to their dissimilar objectives, producing different colour responses that cause brighter colours in the static reconstruction. 
These factors lead to higher quality appearance of the reconstructed model in a static capture. 
\\In this paper we introduce Super-Resolution Appearance Transfer (SRAT) from a human subject acquired with DSLR cameras in a small capture volume to dynamic video performance capture of the same subject acquired with a sparse set of cameras in a larger capture volume.
The objective of SRAT is to enhance the appearance of the 4D reconstruction of a human performance captured in a large volume by exploiting HR images of the same subject acquired with the DSLR cameras. More specifically, the approach improves the colour contrast of the appearance of the 4D reconstruction with a novel colour mapping approach and enhances its fine details by increasing the resolution of the texture maps through a Single-Image Super-Resolution (SISR) network. 
Figure~\ref{fig:firstim} shows how, from the input low-resolution (LR) capture video (Figure~\ref{fig:firstim}a), the final appearance of the 3D model  (Figure~\ref{fig:firstim}c) is 
obtained with appearance detail similar to the HR capture (Figure~\ref{fig:firstim}b).
The contributions presented in this paper are:
\begin{itemize}[noitemsep]
    \item A novel pipeline for super-resolution appearance transfer to enhance the appearance of LR dynamic performances of people from HR images of the same subjects acquired with static DSLR cameras in a small volume. 
    \item A new automatic approach for colour mapping between images acquired with different systems: after the selection of optimal image couples, the colours of the video frames are corrected with an extension of the colour transfer algorithm~\cite{grogan2019l2} to multi-view images.
\end{itemize}
\section{Related Works}
\begin{figure*}[!tbh] 
\centering 
\resizebox{0.9\linewidth}{!}{
\includegraphics[width=1.in]{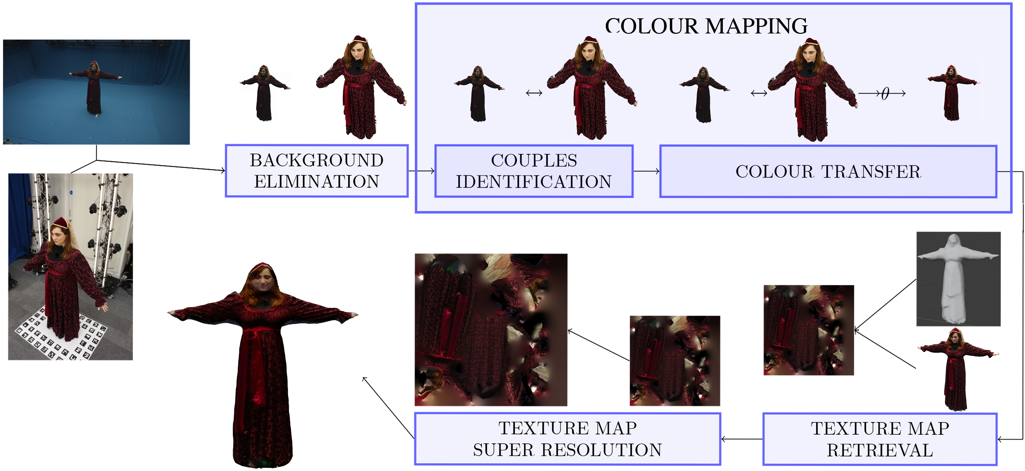}}
\vspace{-1.2em}
\caption{SRAT pipeline: LR capture frames and HR images are the input. After background removal, couple between the most similar frames and HR images are created. The colours of the frames are corrected. Their texture maps are retrieved and finally super-resolved.}
\vspace{-2em}
\label{fig:pipe}
 \end{figure*}
\label{sec:rw}
\textbf{Colour transfer: }colour transfer aims to modify the colours of a target image taking as reference the colours of another image. For some frameworks, there must be pixel correspondences between target and reference images. Examples of these are classic computer vision methods~\cite{finlayson2015color, park2016efficient, hwang2014color, oliveira2014probabilistic}
and deep learning approaches such as image-to-image translation networks~\cite{isola2017image, xian2018texturegan}. Due to the necessity of pixel correspondences, these methods cannot be applied in our case. Early techniques that do not require pixel correspondences define a parametric affine transfer function through statistical moments of colour distributions. These can model only certain types of distributions~\cite{reinhard2001color,pitie2007linear}. Methodologies that use Optimal Transport (OT) framework were then studied but they either introduce grainy artefacts in the gradient of the corrected image~\cite{ferradans2014regularized,pouli2011progressive,freedman2010object} or do not modify the image’s luminance channel~\cite{pitie2007automated,bonneel2015sliced}.
Recent techniques model the colour distribution as Gaussian Mixture models to define correspondences between Gaussian components of the target and reference distributions~\cite{xiang2009selective, jeong2008object, grogan2019l2}. The colour transfer function is learned from a single target-reference image pair, failing in learning a complete map for all the colours of a 3D surface.
Unsupervised cycle-in-cycle neural networks are used to enhance the colours of images as well~\cite{zhu2017unpaired, chen2018deep, he2018deep}. The lack of a consistent amount of training data deteriorates their performances. This paper proposes a colour transfer algorithm that exploits the whole surface of a human model to learn the transfer function from multiple target and reference images without the need of pixel correspondences between them.
\\
\indent \textbf{Single-image super-resolution: }image super-resolution (SR) is an image processing techniques that aims to estimate a perceptually plausible HR image from a LR input image ~\cite{yang2014single}.
Recently, deep neural networks have shown their superior performance on the SISR task. Dong \etal~\cite{dong2014learning} introduced the use of a convolutional neural network (CNN) to super-resolve an image for the first time.
Two main topologies of network architecture were then applied for SR: residual networks~\cite{lim2017enhanced, haris2018deep, li2019feedback, zhang2018image, dai2019second, ahn2019efficient, Mei_2020_CVPR} and generative adversarial networks ~\cite{wang2018esrgan,ledig2017photo,Vasu_2018_ECCV_Workshops,lee2019perceptual,chen2020single}. While the former achieve higher values of peak signal-to-noise ratio (PSNR) and structural similarity index (SSIM) with blurrier outputs, the second have lower figures of the mentioned metrics and unrealistic details in the SR images. 
Reference-based super-resolution (RefSR) is an alternative method that transfers HR textures from a given reference image to super-resolve the LR input~\cite{li2020survey}. CrossNet~\cite{zheng2018crossnet} uses optical flow to align input and reference. To be efficient, the reference and the LR images must have similar content and similar viewpoint. Other approaches~\cite{zhang2019image},\cite{yang2020learning} apply a ``patch-match" mechanism to swap the most similar features of the reference and the LR image. Even though they perform better when the similarity between reference and input image is low, they still introduce unpleasant artefacts. The patch-match mechanism consumes significant amount of GPU memory, making the usage of HR reference images impracticable. 
\\To the best of our knowledge, there are only two deep learning works that aim to super-resolve 3D texture maps of objects. Li \etal~\cite{li20193d} modified the architecture of EDSR~\cite{lim2017enhanced} to exploit the information of both texture maps and normal maps of objects. Richard \etal~\cite{richard2019learned} combined a redundancy-based part with a prior-based part in a network to create new texture maps. The first method requires the creation of normal maps, a process that introduces heavy computational cost for a high number of frames. The second approach is mainly oriented in the creation of texture maps. 
Following these two methods, we use a residual neural network in SRAT that, differently from 2D SISR networks, is trained with datasets of texture maps of people.
\section{Methodology}
\noindent
SRAT aims to improve the appearance of LR capture video of human performance acquired with a sparse multi camera system by leveraging a collection of HR images of the same subject. The sparse multi camera system allows dynamic capture of human performance that covers a large volume at the cost of reduced resolution on the subject. In contrast, the static HR images are captured using an array of DSLR cameras resulting in increased resolution on the subject including fine details in the hair, skin and clothing.
To globally improve the appearance of the performance, we first tackle the problem that these systems will have a different capture settings and colour response by automatically learning a colour transfer function between the two systems (Section~\ref{ssec:colormap}). The fine details of the dynamic LR texture maps are then enhanced by leveraging a learnt SR model and thus, the appearance is locally improved (Section~\ref{ssec:super}).  
\subsection{Overview}
\label{ssec:over}
\noindent
The input data of the proposed approach are:
\begin{itemize}[noitemsep]
    \item LR capture video with $N_{LR}$ camera. This consists of: RGB image set of a subject $\{\{I^i_{LR_l}(t)\}_{l=1}^{N_F}\}_{i=1}^{N_{LR}}$; $N_F$ 3D meshes (one for every frame); $N_F$ texture maps;
    \item HR image set of the same subject captured with a rig of $N_{HR}$ DSLR cameras $\{I^j_{HR}\}_{j=1}^{N_{HR}}$
\end{itemize}
Without loss of generality, $I_{LR}$ is a time instant of $I_{LR}(t)$.
SRAT consists of 4 stages shown in Figure~\ref{fig:pipe} and outlined below.
\\\textbf{Background elimination: }to ensure an accurate result of the colour mapping, the background of the frames and HR images must be removed. For the former, we use silhouettes computed via Chroma keying while for the latter the alpha matting method proposed by Hu and Clark~\cite{hu2019instance} is applied.
\\\textbf{Colour mapping: }the second stage of the pipeline is a novel approach to map the colour features between images acquired with different systems. It consists of two steps. (i) \textit{Couple identification: }each $I_{LR}$ is paired with one $I_{HR}$ to create couple of similar images by performing the similarity evaluation among partial texture maps. (ii) \textit{Colour transfer: }a colour transfer function is learned from the multiple couples of the previous step to correct the colours of all $I_{LR}$.
\\\textbf{Texture map retrieval: }texture maps are retrieved by projecting the new corrected frames to the corresponding meshes reconstructed in a pre-processing step.
\\\textbf{Texture map SR: }the details of the texture maps are enhanced by super resolving them with an RCAN-style network~\cite{zhang2018image} trained with human texture maps.
\\We make 3 important assumptions that affect the design of the pipeline and its evaluation: (i) geometric properties of the HR static model are not exploited to avoid estimation of geometric surface correspondence between the LR and HR reconstructions; (ii) the color response of the cameras within each capture system is the same; (iii) in the selection of a subset of $n<N_{LR}$ cameras for evaluation we assume evenly spaced cameras to maximise coverage of the subject. 
\subsection{Colour mapping of frame images}
\label{ssec:colormap}
\noindent
The camera responses of the two capture systems differ due to their environment settings that are defined based on the specific purposes of the captures. The contrast range of the DSLR cameras allows to acquire HR images with brighter colours and higher resolution. To obtain the same contrast in the LR capture video, the colours of the HR images are mapped to its frames with a novel approach that comprises 2 steps: pairs between the most similar images of the two systems are created according to surface visibility and viewing angle pairs (couples identification); colours of the LR capture frames are corrected with a colour transfer function estimated from multi-view images (colour transfer).
\\\indent\textbf{Couples identification: }the algorithm that learns the colour transfer function to correct the colours of the frame images, requires a pair of target and reference images as input. In our case, the target images are the LR capture frames while the reference images are the HR images. Since a colour palette is defined from the reference images, the algorithm performs better if the content of $I_{LR}$ is similar to the content of $I_{HR}$. If for instance, a reference image that shows the back of the person has been paired with a target frame representing the front part of the subject, the algorithm may fail to transfer the colour of the face because it is not learned from the visible colour palette (Section~\ref{ssec:ctres}). In the two systems, the camera settings are different, resulting in different representations of the same model. Creating the pairs by directly comparing the original images is ineffective and thus, we operate in the texture map domain. We propose a new automatic method to identify the couples between images of different systems. Since no geometric information of the HR model is available, we reconstruct partial texture maps of the frames $\{\{T(I^i_{LR_l})\}_{l=1}^{N_F}\}_{i=1}^{N_{LR}}$ and of the HR images $\{T(I^j_{HR})\}_{j=1}^{N_{HR}}$ by applying Densepose~\cite{alp2018densepose} and unwrapping the resulting UV maps. 
These partial texture maps are invariant to the camera orientation and position allowing a comparison between the two systems in a common domain. We evaluate the similarity between all the $T(I^i_{LR_l})$ and $T(I^j_{HR})$ with the SSIM metric~\cite{1284395}. Each $I_{LR}$ is coupled with one $I_{HR}$ whose partial texture map is the most similar defined by the SSIM metric as shown in Equation~\ref{eq:1}:  
\begin{equation}
\label{eq:1}
\displaystyle
I^i_{LR_l} \leftrightarrow I^j_{HR}\; where\; \argmax_{I^j_{HR}}\{SSIM(T(I^i_{LR_l}),T(I^j_{HR}))\} 
\end{equation}
where $I^i_{LR_l}$ is the $l^{th}$ frame of the $i^{th}$ camera and $I^j_{HR}$ is the HR image of the $j^{th}$ camera.
%
\\
\indent\textbf{Colour transfer: }
the idea of the colour transfer algorithm applied is based on ~\cite{grogan2019l2}. In their work, the parameters of the colour transfer function are learned from a single pair of target and reference images. In our case, the function must learn a map from all the colours of the static acquisition to the colours of $I_{LR}$: all the view-angles of the human model must be seen during the learning stage. We therefore extend the cited work~\cite{grogan2019l2} to more than two images as inputs. We model the colour transfer function $\phi_{\theta}(x)$ as a Thin Plate Splines function that depends on a set of parameters $\theta$. This set is computed by minimising the following novel energy function with gradient descent algorithm:
\begin{equation}
\label{eq:2}
\theta=\frac{1}{N}\sum_{l=1}^{N}\argmin_{\theta_l}\{||p_f||^2-2<p_f|p_I>\}\tag{2}
\end{equation}
where $N$ is the number of input couples, 
$p_f$ is the distribution of $I_{LR}$ with parameterised mean $\phi_\theta(\mu_f)$ and $p_I$ is the distribution of $I_{HR}$. Equation~\ref{eq:2} is further explained in the supplementary material.
For each selected input couple, a set of parameters is computed by minimising the energy function. The parameters of $\phi_{\theta}(x)$ are obtained averaging these sets and used to correct the colours of all the frames.
\subsection{Texture map super-resolution}
\label{ssec:super}
\begin{table*}[!h]
\centering
\resizebox{0.85\linewidth}{!}{\input{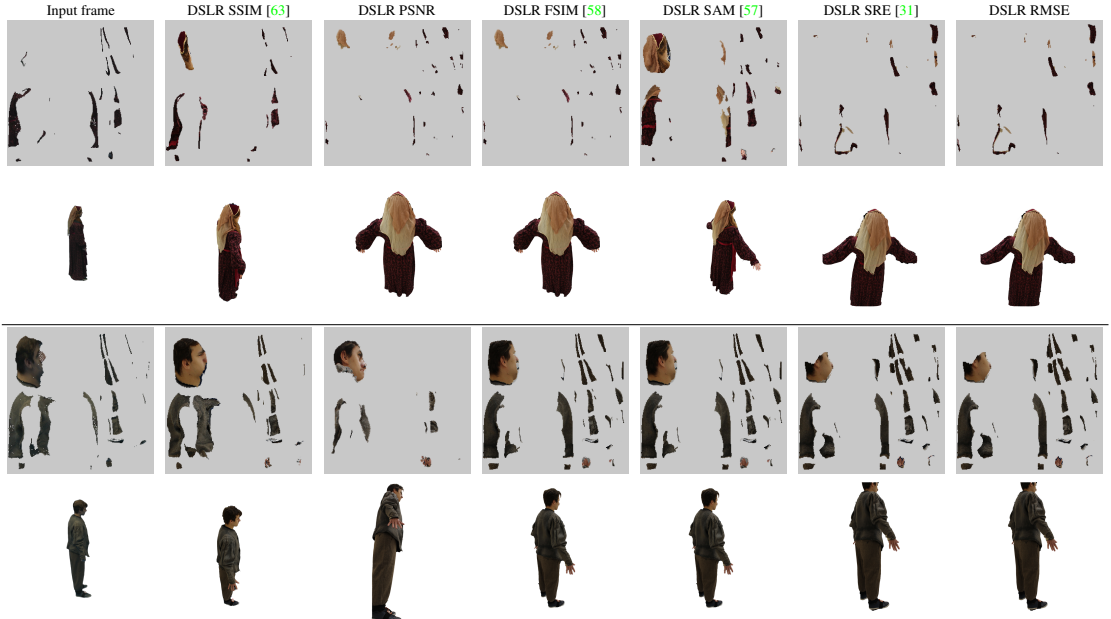}}
 \vspace{-1.2em}
  \captionof{figure}{The first column shows the partial texture map and the LR capture frame for \textit{SingleF} (first two rows) and \textit{SingleM} (last two rows). The images and partial texture maps of the other columns are acquired with the DLSR system and paired with different similarity metrics.}
  \vspace{-1.2em}
 \label{fig:coup1}
\end{table*}
\noindent
After having retrieved the texture maps from the corrected frames, we treat them as 2D RGB images and we give them as input to the RCAN-style network~\cite{zhang2018image}. The residual-in-residual RCAN network super-resolves the obtained texture maps through a channel attention mechanism that adaptively rescale each channel-wise feature by modelling the interdependencies across feature channels.
In its original work, RCAN was trained augmenting 800 RGB images from DIV2K dataset~\cite{Agustsson_2017_CVPR_Workshops}. Since we aim to super-resolve texture maps of a specific model, we first pre-train RCAN with a set of patches of human texture maps~\cite{renderpeople}. We then fine-tune it with the $N_F$ original texture maps of the input model. Further details of its training are presented in the supplementary material. 
We finally assign the SR texture maps to the correspondent reconstructed meshes to obtain the enhanced 4D reconstruction of the performance.
\section{Results evaluation}
\label{sec:results}
\noindent
We capture two subjects, one male (\textit{SingleM}) and one female (\textit{SingleF}), using both LR capture video acquired with $N_{LR}=16$ cameras for $N_F=440, 470$ frames respectively and a subset of $N_{HR}=64$ DSLR cameras. We evaluate the proposed approach on the 4D performances of the models reconstructed with the method proposed by Starck and Hilton~\cite{starck2007surface}, applied in the pre-processing step to retrieve the input data. Additional visual results and evaluations are presented in the supplementary material.
\subsection{Couples identification}
\label{ssec:srmeth}
\begin{table*}[!h]
    \centering
     \resizebox{0.9\linewidth}{!}{\input{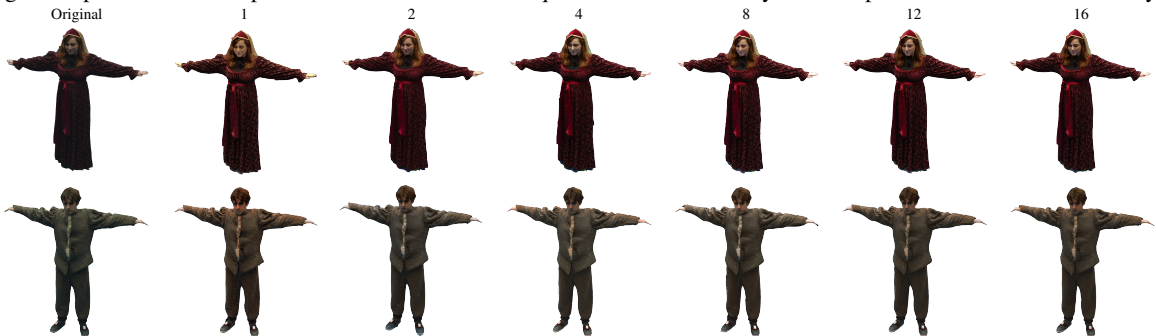}}   
     \vspace{-1.2em}
     \captionof{figure}{Colour transfer with different numbers of input couples. In the top row \textit{SingleF} model, in the bottom \textit{SingleM}.}
     \vspace{-2em}
     \label{fig:cc1}
    \end{table*}
\begin{table}[!]
\centering
\resizebox{0.85\linewidth}{!}{\begin{tabular}{|c|cc|cc|}
\hline
\multicolumn{1}{|c|}{\multirow{2}{*}{\textbf{Input couples}}} & \multicolumn{2}{c|}{\textbf{SingleF}}                            & \multicolumn{2}{c|}{\textbf{SingleM}}                            \\ \cline{2-5} 
\multicolumn{1}{|c}{}                         & \multicolumn{1}{|c}{JS $\downarrow$} & \multicolumn{1}{|c}{$\chi^2\downarrow$} & \multicolumn{1}{|c}{JS $\downarrow$} & \multicolumn{1}{|c|}{$\chi^2\downarrow$   }\\ \hline

1                                                                                              & 0.5452              & 18.8659                       & 0.4759                  & 21.5098                       \\
2                                                                                              & 0.5439                  & 14.7575                       & 0.4691                  & 18.3891                       \\
4                                                                                              & 0.5435                  & 14.9826                       & 0.4649                  & 22.1790                       \\
8                                                                                              & 0.5423                 & 13.5905                       & 0.4609                  & 18.0991                       \\
12                                                                                             & {\textbf{0.5321}}
& {\textbf{12.9507}}                       & {\textbf{0.4542}}                  & {\textbf{15.5898}}                      \\
16                                                                                             &    0.5443                  & 14.0894                       & 0.4624                  & 21.3688                    \\ \hline
\end{tabular}}
\vspace{-0.8em}
\caption{JS and $\chi^2$ values for different numbers of input couples. $\downarrow$ indicates lower value is better.}
\vspace{-2.2em}
\label{tbl:cc1}
\end{table}
\noindent
The first study was conducted on the first part of the colour mapping stage. We evaluate the effect of using different similarity metrics to pair the partial texture maps of the two acquisition systems. We compute the similarity with the following metrics: SSIM~\cite{1284395}, PSNR, feature based similarity index (FSIM~\cite{zhang2011fsim}), spectral angle mapper (SAM~\cite{yuhas1992discrimination}), signal to reconstruction error ratio (SRE~\cite{lanaras2018super}), root mean square error (RMSE). Figure~\ref{fig:coup1} shows the partial texture maps of the HR images associated with the partial texture map of the LR capture frames and the correspondent original images for each of the studied metrics. In the case of \textit{SingleF}, most of the metrics create a pair with two images where the pose of the subject is significantly different. SSIM associates the most similar HR image to the input frame. The orientation of \textit{SingleM} in the HR images paired with other metrics do not match the one in the input frame. In these HR images, the nose of the subject is hardly visible oppositely from the HR image paired with SSIM.
\subsection{Colour transfer input couples selection}
\noindent
We analyse the effect of using different numbers of input couples in the colour transfer step. We expect the algorithm to perform better if a complete coverage of the surface is exploited in the learning of the colour transfer function. The effect of having 1, 2, 4, 8, 12 and 16 couples as input is evaluated. The $N_{LR}$ cameras are equally distributed in a studio, with 4 cameras on each side of a square surrounding the acquisition space. In the case of 12 couples, 3 cameras from each side are selected; in the case of 8, 2 cameras each side; in the case of 4, just 1 camera each side. 
With less input couples (1 and 2), there is a lower surface coverage and the model is not completely captured.
As quantitative evaluation, the dissimilarity between a specific corrected frame and its paired HR image is computed considering their colour histograms since the pixel-based metrics are not effective due to the different nature of the frame and the HR image. We compute the Jensen-Shannon (JS) divergence and the Chi-Squared ($\chi^2$) distance between the normalized colour histograms of all the corrected frames and of the correspondent HR images. 
Even though these two metrics have been revealed to be the most efficient~\cite{zhang2014comparison}, they present some drawbacks. JS focuses only on the statistical distribution of the images while $\chi^2$ only accounts for the difference between the corresponding bins and is hence sensitive to distortions and quantization~\cite{yang2015chi}.
The average of the distances for every corrected couple is presented in Table~\ref{tbl:cc1}. Generally, the JS divergence decreases when the number of couples increases for both the datasets (except for the case with 16 couples). 
The $\chi^2$ distance follows the same pattern except for the 4 couple case because of possible distortions of the corrected frames. As expected, the worst results are given by the 1 couple case. As shown in Figure~\ref{fig:cc1}, if 1 couple is the input, the face of the models have unnatural colours. The algorithm did not learn how to transfer the face colour as it was not seen in the input images during learning. For the same reason, the outputs of the 2 couple case seem blurry. Even though the lowest figures are given by having 12 couples as input, the results produced with 8 couples are quantitatively and qualitatively satisfactory and we use them throughout the presented results balancing visual quality with computational complexity of SRAT.
\begin{table}
\centering
\resizebox{0.85\linewidth}{!}{\begin{Large}

\begin{tabular}{cc}

\includegraphics[width=1.3\linewidth]{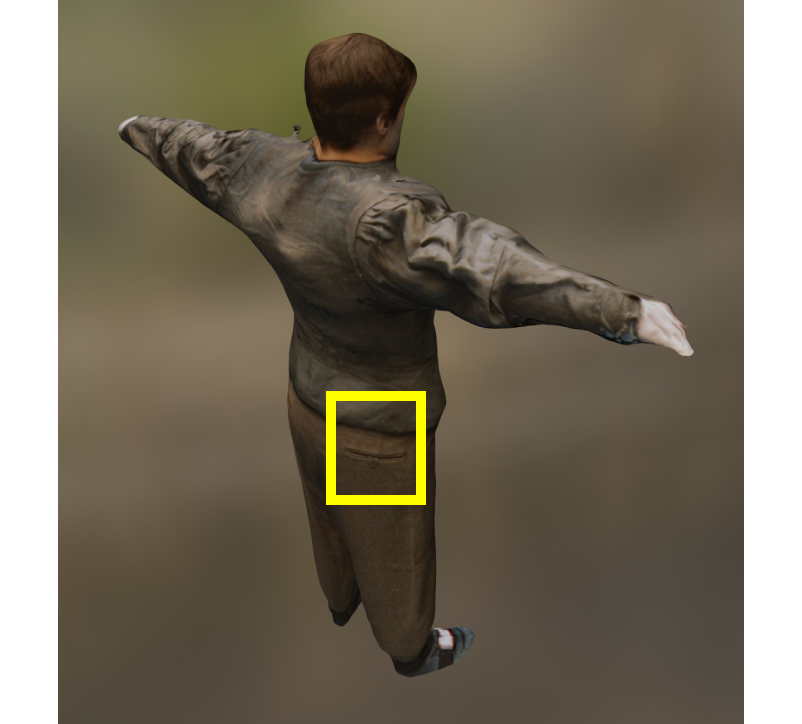} &
\includegraphics[width=1.2\linewidth]{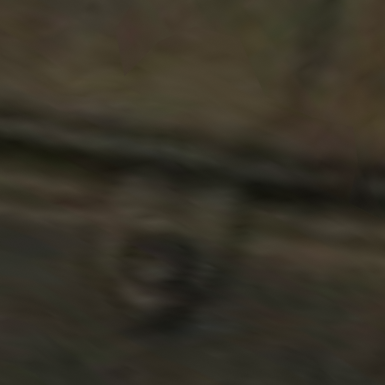} \\ 
(a) Output model & (b) Original LR\\
\includegraphics[width=1.2\linewidth]{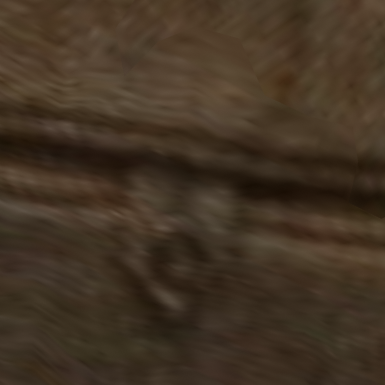}  &
\includegraphics[width=1.2\linewidth]{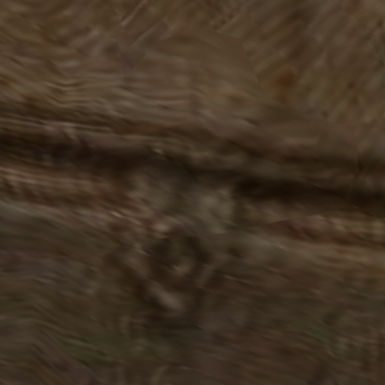}  \\
(c) Corrected LR & (d) 2x SR
\end{tabular}
\end{Large}}
\vspace{-1.0em}
\captionof{figure}{Detail of \textit{SingleM} model rendered with the original (b), the corrected (c) and the super-resolved (d) texture map.}
\vspace{-1.em}
\label{fig:supout2}
\end{table}
 \begin{table}[!]
\centering
\resizebox{0.9\linewidth}{!}{\begin{tabular}{|c|c|cc|cc|}
\hline
\multicolumn{1}{|l|}{\multirow{2}{*}{\textbf{Models}}} & \multicolumn{1}{l|}{\multirow{2}{*}{\textbf{Scale}}} & \multicolumn{2}{c|}{\textbf{SingleF}}                          & \multicolumn{2}{c|}{\textbf{SingleM}}                          \\ \cline{3-6} 
\multicolumn{1}{|l|}{}                                & \multicolumn{1}{l|}{}                       & \multicolumn{1}{l|}{PSNR $\uparrow$} & \multicolumn{1}{l|}{SSIM $\uparrow$} & \multicolumn{1}{l|}{PSNR $\uparrow$} & \multicolumn{1}{l|}{SSIM $\uparrow$} \\ \hline
1a                                                    & 2x                                          & 48.53                     & 0.9910                    & 50.31                     & 0.9935                    \\
1b                                                    & 2x                                          & {\textbf{48.64}}      & {\textbf{0.9912}}     & {\textbf{50.45}}      & { \textbf{0.9937}}     \\
1c                                                    & 2x                                          & 48.60                     & 0.9910                    & 50.41                     & 0.9937                    \\
2a                                                    & 2x                                          & 47.68                     & 0.9892                    &    48.91                       &      0.9914                     \\
2b                                                    & 2x                                          & 47.70                     & 0.9893                    &     48.94                      &        0.9915                   \\
2c                                                    & 2x                                          & 47.77                     & 0.9895                    &      40.01                     &             0.9916              \\
3a                                                    & 2x                                          & 47.74                     & 0.9893                    &        48.92                   &           0.9915                \\
3b                                                    & 2x                                          & 47.85                    & 0.9895                    &       49.22                    &         0.9918                  \\
3c                                                    & 2x                                          & 28.68                     & 0.8737                    &         49.33                  &          0.0019                 \\
                                                                                
\hline


\end{tabular}

}
\vspace{-0.8em}
\caption{PSNR and SSIM results of different training configurations of RCAN. $\uparrow$ indicates higher value is better.}
\vspace{-2em}
\label{tbl:SR1}
\end{table}
\label{ssec:ctres}
\begin{table*}[h!]
    \centering
       \resizebox{0.85\linewidth}{!}{\input{srat/image_files/pipelines}}
     \vspace{-1.2em}
     \captionof{figure}{Visual results of different configurations of SRAT for \textit{SingleF} (top) and \textit{SingleM} (bottom)  models.}
     \vspace{-1.0em}
     \label{fig:pipelines}
    \end{table*}
\subsection{Training models}
\noindent
We analyse the effect of using different trained models for RCAN. We train the network with three datasets: 
\begin{enumerate}[noitemsep]
    \item Original texture maps of the LR input model.
    \item Original HR images of the model acquired with the DSLR system.
    \item Same as 2 but with a removal of the background.
\end{enumerate}
and in three configurations:
\begin{enumerate}[label=(\alph*),noitemsep]
    \item RCAN is trained with only our datasets.
    \item RCAN is trained with a dataset made of texture maps of 7 human models~\cite{renderpeople} cropped into patches and fine-tuned with our datasets.
    \item RCAN is trained with the DIV2K dataset as in the original paper and fine-tuned with our datasets.
\end{enumerate}
\begin{table*}[!]
    \centering
        \resizebox{0.85\linewidth}{!}{\input{srat/image_files/cc2}}
     \vspace{-1.0em}
     \captionof{figure}{Outputs of different colour transfer approaches. In the top row \textit{SingleF} model, in the bottom \textit{SingleM}.}
     \vspace{-1.8em}
     \label{fig:cc2}
    \end{table*}
\begin{table}[!]
\centering
 \resizebox{0.8\linewidth}{!}{\begin{tabular}{|c|cc|cc|}
\hline
\multicolumn{1}{|c|}{\multirow{2}{*}{\textbf{Methods}}} & \multicolumn{2}{c|}{\textbf{SingleF}}                            & \multicolumn{2}{c|}{\textbf{SingleM}}                            \\ \cline{2-5} 
\multicolumn{1}{|c}{}                         & \multicolumn{1}{|c}{JS $\downarrow$} & \multicolumn{1}{|c}{$\chi^2\downarrow$} & \multicolumn{1}{|c}{JS $\downarrow$} & \multicolumn{1}{|c|}{$\chi^2\downarrow$} \\ \hline
Vander~\cite{klinger1967vandermonde}                                    & 0.8323                  & \textbf{1.90568}                       & 0.8142                  & 48.6026                       \\
Finlayson~\cite{finlayson2015color}                                      & 0.7949                  & 50.4117                       & 0.6375                  & 121.647                       \\
Pitie~\cite{pitie2007linear}                                            &         0.7886         &    38.9494                          &             0.7445             &                    88.2087                  \\
CycleGan~\cite{zhu2017unpaired}                                      &             0.5776             &            584.419                   &             0.6158            &          336.151                 \\
$TPS$~\cite{grogan2019l2}                                      & \textbf{0.5029}                  & 120.177                       & \textbf{0.4479}                  & 191.874                       \\
SRAT(ours)                                           & 0.5435                  & 13.5905                       & 0.4609                  & \textbf{18.0991}                      
\\ \hline
\end{tabular}
}
\vspace{-0.9em}
\caption{JS and $\chi^2$ values for different colour transfer algorithms.}
\vspace{-2.3em}
\label{tbl:cc2}
\end{table}
The texture maps retrieved in the 3rd stage with the methods proposed by Allene \etal~\cite{allene2008seamless} are bicubic downscaled ($2\times$) and then super-resolved with the different trained models of RCAN. PSNR and SSIM values are computed between SR and HR texture maps and presented in Table~\ref{tbl:SR1}. For both \textit{SingleM} and \textit{SingleF}, the 1b model achieves the highest values. Visual results are presented in the supplementary material.
Figure~\ref{fig:supout2} shows a detail of \textit{SingleM} model with the corrected SR texture map. Compared to the original model, the button looks sharper and less blurry, showing the efficiency of SR application to enhance fine details of the appearance.
\subsection{Configuration of pipeline}
\label{ssec:config}
\noindent
Another evaluation is performed by changing the order of the stages in SRAT. The first stage and the first step of the second stage are not modified. The configurations are:
\begin{enumerate}[noitemsep]
    \item (i) Texture map retrieval; (ii) Texture map colour transfer; (iii) Texture map SR. 
    \item (i) Texture map retrieval; (ii) Texture map SR; (iii) Texture map colour transfer. 
    \item (i) Input frames colour transfer; (ii) Corrected frames SR; (iii) Texture map retrieval. 
    \item (i) Input frames SR; (ii) Input frames colour transfer; (iii) Texture map retrieval. 
    \item (i) Input frames SR; (ii) Texture map retrieval; (iii) Texture map colour transfer. 
\end{enumerate}
We apply the output texture maps to the 3D models and only a qualitative evaluation is performed due to the lack of a ground-truth. Figure~\ref{fig:pipelines} shows \textit{SingleF} and \textit{SingleM} models for each configuration. Compared to the original model, the appearance is improved, with brighter colours and more visible details. If the colour transfer is done after the texture map retrieval (configurations 1 and 2), the dress presents artefacts (3rd and 6th rows). 
If the SR stage is applied before the texture map retrieval (configurations 3, 4 and 5), noise is introduced as seen on the face (2nd and 5th rows). 
\\\indent\textbf{Performance evaluation:} the SR stage requires more time ($\sim128s/197s$ per texture map/frame) than the colour transfer for LR images ($\sim8s/14s$) and for SR images ($\sim35s/56s$).
The 1st configuration is the fastest and least computationally expensive: the colour transfer and the SR stages are applied to the texture maps, which are less in number than the frames. The slowest and the most computationally expensive configuration is the 4th one because the two stages are applied to the input frame (16 images of the video cameras for every frame).
\begin{table*}[!]
    \centering
        \resizebox{0.85\linewidth}{!}{\input{srat/image_files/SR2}}
     \vspace{-1.1em}
     \captionof{figure}{Visual comparison of $2\times$ SR networks. At the bottom, heatmaps of the above \textit{SingleF} texture map portion.}
     \vspace{-1.9em}
     \label{fig:SR2}
    \end{table*}
\begin{table}[!]
 \centering
\resizebox{0.85\linewidth}{!}{\begin{tabular}{c|c|cc|cc|}
 \cline{2-6} 
& \multirow{2}{*}{\textbf{Methods}}                  & \multicolumn{2}{c|}{\textbf{SingleF}}                                                & \multicolumn{2}{c|}{\textbf{SingleM}}                                                \\ \cline{3-6} 
                                            &       & \multicolumn{1}{c|}{PSNR $\uparrow$} & SSIM $\uparrow$                  & \multicolumn{1}{c|}{PSNR $\uparrow$} & SSIM $\uparrow$                  \\ \hline
\multicolumn{1}{|c|}{\multirow{9}{*}{\rotatebox[origin=c]{90}{\textbf{\textit{SISR $\bf 2\times$}}}}} &
Bicubic                                            & 45.96                                             & 0.9807                           & 47.39                                             & 0.9885                           \\
\multicolumn{1}{|c|}{}                        &
DBPN~\cite{haris2018deep}    & 20.91                                             & 0.3000                           & 19.73                                             & 0.1188                           \\
\multicolumn{1}{|c|}{}                        &
SRFBN~\cite{li2019feedback}  & 17.12                                             & 0.7583                           & 17.67                                             & 0.7381                           \\
\multicolumn{1}{|c|}{}                        &
CSNLN~\cite{Mei_2020_CVPR} & 48.12                                             & 0.9900                           & 49.36                                             & 0.9920                           \\
\multicolumn{1}{|c|}{}                        &
NLR~\cite{li20193d}          & 44.95                                             & 0.9856                           & 46.73                                             & 0.9888                           \\
\multicolumn{1}{|c|}{}                        &
NHR~\cite{li20193d}          & 34.81                                             & 0.9788                           & 42.48                                             & 0.9862                           \\
\multicolumn{1}{|c|}{}                        &
RCAN~\cite{zhang2018image}   & 48.14                                             & 0.9900                           & 49.35                                             & 0.9919                           \\
\multicolumn{1}{|c|}{}                        &
RCANtext                                           & 47.51                                             & 0.9886                           & 48.50                                             & 0.9904                           \\
\multicolumn{1}{|c|}{}                        &
SRAT(ours)                         & \textbf{48.64}                   & \textbf{0.9912} & \textbf{50.45}                   & \textbf{0.9937} \\ \hline
\multicolumn{1}{|c|}{\multirow{4}{*}{\rotatebox[origin=c]{90}{\textbf{\textit{RefSR $ \bf 4\times$}}}}} &
CrossNet~\cite{zheng2018crossnet}                                          &     37.16                                              &                 0.9155                  & 38.96                                             & 0.9438                           \\
\multicolumn{1}{|c|}{}                        &
TTSR~\cite{yang2020learning}                                               & 41.29                                             & 0.9427                           & 43.00                                             & 0.9648                           \\
\multicolumn{1}{|c|}{}                        &
TTSR-\textit{$l_2$}               & 42.86                                             & 0.9622                           & 44.75                                             & 0.9759                           \\
\multicolumn{1}{|c|}{}                        &
SRNTT~\cite{zhang2019image}                                               & 42.92                                             & 0.9542                           & 43.86                                             & 0.9677                           \\
\multicolumn{1}{|c|}{}                        &
SRNTT-\textit{$l_2$}              & 44.11                                             & 0.9626                           & 45.26                                             & 0.9760                           \\ 
\hline
\multicolumn{1}{|c|}{}                        &
SRAT($4\times$)                         & \textbf{44.42}                   & \textbf{0.9636} & \textbf{45.57}                   & \textbf{0.9770} \\\hline
\end{tabular}}
\vspace{-1.0em}
\caption{Quantitative results of different SR approaches.}
\vspace{-2.em}
\label{tbl:SR2}
\end{table}
\subsection{Comparison with related works}
\label{ssec:relcomp}
\noindent
We compare the colour transfer algorithm and the SR network used in SRAT with related works.
\\\indent\textbf{Colour transfer approaches: }the JS and $\chi^2$ values of 5 colour transfer frameworks are shown in Table~\ref{tbl:cc2}. 
Finlayson~\cite{finlayson2015color} and Vander~\cite{klinger1967vandermonde} require pixel correspondences. Pitie~\cite{pitie2007linear} introduces a parametric colour transfer function and CycleGan~\cite{zhu2017unpaired} is an unsupervised deep learning technique trained with our frames and HR images. The last method is the framework TPS~\cite{grogan2019l2} without any modifications. As shown in Figure~\ref{fig:cc2}, Vander, Finlayson and CycleGan cannot transfer the colours from the HR image to the frames while Pitie outputs are too bright. If TPS is applied, a different function for every frame of each camera is learned. Therefore, the corrected frames of the same scene and the ones of two consecutive frames have different colours. W.r.t. the quantitative analysis, the JS divergence is the lowest when TPS is applied. TPS learns the colour transfer function by modelling the statistical distributions of the input images. 
The JS divergence focuses on the difference between statistical distributions and it is lower if these distributions are learned for every frame and not only for 8 selected couples as in our case. The $\chi^2$ of TPS is the second highest. Its lowest figure is obtained when Vander is applied for \textit{SingleF} and ours for \textit{SingleM}. Vander colour transfer produces wrong outputs and the low $\chi^2$ value is influenced by its sensitiveness to image distortion.
\\\textbf{SR approaches: }the top part of Table~\ref{tbl:SR2} shows quantitative comparisons for $2\times$ SR with a classic computer vision approach (Bicubic) and 7 deep learning methods: DBPN~\cite{haris2018deep}, SRFBN~\cite{li2019feedback}, CSNLN~\cite{Mei_2020_CVPR}, NLR and NHR~\cite{li20193d}, RCAN~\cite{zhang2018image} and RCANtext. For testing with NLR and NHR, the normal maps were retrieved with Blender2.8~\cite{blender}. RCAN is the original version and RCANtext is RCAN trained with the human texture map dataset but not fine-tuned with the original texture maps as done in SRAT, which outperforms all the tested methods for both the datasets. The four best network outputs and the heatmaps of the difference with the ground-truth are shown in Figure~\ref{fig:SR2}: SRAT heatmap presents more blue pixels confirming its superiority. We then compare SRAT with 3 state-of-the-art RefSR methods ($4\times$ super-resolution). Since no training dataset of texture maps with relative references has been available online, we train Cross-net~\cite{zheng2018crossnet}, TTSR~\cite{yang2020learning} and SRNTT~\cite{zhang2019image} with CUFED5~\cite{zhang2019image} dataset as done in their papers. During inference, we select a HR image for each model as reference and we downscale it ($6\times$) for the problem of GPU memory consumption. For the same reason, the LR texture maps are cropped into patches (64x64 size). We train SRNTT and TTSR with only the reconstruction loss (indicated with the suffix \textit{$l_2$}) for a fair comparison with SRAT. In the bottom part of Table~\ref{tbl:SR2}, the PSNR and SSIM figures confirm the superiority of SRAT to RefSR methods.
\subsection{More complex scenarios}
\begin{table}[!]
  \centering
        \resizebox{0.85\linewidth}{!}{\input{srat/image_files/poses}}
    \vspace{-1.0em}
     \captionof{figure}{Complex scenarios with multiple performers captured simultaneously (a,b) and unseen poses (c,d)}
     \vspace{-2.1em}
\label{fig:poses}
\end{table}
\label{ssec:complex}
\textbf{Interaction scenario: }we apply SRAT to an interaction scene where \textit{SingleF} and \textit{SingleM} perform at the same time. A colour transfer function is learned for each model using the same couples of SRAT as input. The colours of the frames are then corrected by applying the two functions to their correspondent models selected with a mask. For each model, its texture maps are retrieved for each frame and super-resolved with the two fine-tuned SR networks. 
\\\indent\textbf{Unseen Poses: }the proposed pipeline aims to enhance the appearance of a human perfomance. Therefore, it has to handle multiple poses of the performers. SRAT is effective also when it is applied to frames that were not used to learn either the colour transfer function or the SR model.
\\Figure~\ref{fig:poses} shows the results of these complex scenarios.
\section{Conclusion}
\label{sec:conclusion}
\noindent
This paper proposes a novel pipeline to enhance the dynamic appearance of low-resolution capture video of human performance using a collection of static high-resolution images of the same subject. The pipeline enables multi-view performance capture systems to increase the capture volume without sacrificing the output reconstruction quality. A novel automatic colour mapping improves the global appearance by correcting the colours of LR capture frames while fine-scale surface details are transferred by an RCAN-style network from the high-resolution images to the super-resolved texture maps. A limitation of the proposed pipeline is that it does not enforce any temporal coherence between the super-resolved texture maps of consecutive frames. This as well as geometric detail transfer between the models of the two systems will be investigated as future works.

{\small
\bibliographystyle{ieee_fullname}
\bibliography{egbib}

\begin{thebibliography}{10}\itemsep=-1pt

\bibitem{blender}
Blender.
\newblock https://www.blender.org/.
\newblock Accessed: 2020-07-26.

\bibitem{renderpeople}
Renderpeople.
\newblock https://renderpeople.com/.
\newblock Accessed: 2020-07-26.

\bibitem{Agustsson_2017_CVPR_Workshops}
Eirikur Agustsson and Radu Timofte.
\newblock Ntire 2017 challenge on single image super-resolution: Dataset and
  study.
\newblock In {\em The IEEE Conference on Computer Vision and Pattern
  Recognition (CVPR) Workshops}, July 2017.

\bibitem{ahn2019efficient}
Namhyuk Ahn, Byungkon Kang, and Kyung-Ah Sohn.
\newblock Efficient deep neural network for photo-realistic image
  super-resolution.
\newblock {\em arXiv preprint arXiv:1903.02240}, 2019.

\bibitem{allene2008seamless}
C{\'e}dric All{\`e}ne, Jean-Philippe Pons, and Renaud Keriven.
\newblock Seamless image-based texture atlases using multi-band blending.
\newblock In {\em 2008 19th International Conference on Pattern Recognition},
  pages 1--4. IEEE, 2008.

\bibitem{alp2018densepose}
R{\i}za Alp~G{\"u}ler, Natalia Neverova, and Iasonas Kokkinos.
\newblock Densepose: Dense human pose estimation in the wild.
\newblock In {\em Proceedings of the IEEE Conference on Computer Vision and
  Pattern Recognition}, pages 7297--7306, 2018.

\bibitem{bhatnagar2019multi}
Bharat~Lal Bhatnagar, Garvita Tiwari, Christian Theobalt, and Gerard Pons-Moll.
\newblock Multi-garment net: Learning to dress 3d people from images.
\newblock In {\em Proceedings of the IEEE International Conference on Computer
  Vision}, pages 5420--5430, 2019.

\bibitem{Bi_2020_CVPR}
Sai Bi, Zexiang Xu, Kalyan Sunkavalli, David Kriegman, and Ravi Ramamoorthi.
\newblock Deep 3d capture: Geometry and reflectance from sparse multi-view
  images.
\newblock In {\em Proceedings of the IEEE/CVF Conference on Computer Vision and
  Pattern Recognition (CVPR)}, June 2020.

\bibitem{bonneel2015sliced}
Nicolas Bonneel, Julien Rabin, Gabriel Peyr{\'e}, and Hanspeter Pfister.
\newblock Sliced and radon wasserstein barycenters of measures.
\newblock {\em Journal of Mathematical Imaging and Vision}, 51(1):22--45, 2015.

\bibitem{chen2018deep}
Yu-Sheng Chen, Yu-Ching Wang, Man-Hsin Kao, and Yung-Yu Chuang.
\newblock Deep photo enhancer: Unpaired learning for image enhancement from
  photographs with gans.
\newblock In {\em Proceedings of the IEEE Conference on Computer Vision and
  Pattern Recognition}, pages 6306--6314, 2018.

\bibitem{chen2020single}
Zhiyong Chen, Jing Hu, Xuyang Zhang, and Xiangjun Li.
\newblock Single image super-resolution based on enhanced deep residual gan.
\newblock In {\em MIPPR 2019: Pattern Recognition and Computer Vision}, volume
  11430, page 114301W. International Society for Optics and Photonics, 2020.

\bibitem{collet2015high}
Alvaro Collet, Ming Chuang, Pat Sweeney, Don Gillett, Dennis Evseev, David
  Calabrese, Hugues Hoppe, Adam Kirk, and Steve Sullivan.
\newblock High-quality streamable free-viewpoint video.
\newblock {\em ACM Transactions on Graphics (ToG)}, 34(4):1--13, 2015.

\bibitem{dai2019second}
Tao Dai, Jianrui Cai, Yongbing Zhang, Shu-Tao Xia, and Lei Zhang.
\newblock Second-order attention network for single image super-resolution.
\newblock In {\em Proceedings of the IEEE conference on computer vision and
  pattern recognition}, pages 11065--11074, 2019.

\bibitem{de2008performance}
Edilson De~Aguiar, Carsten Stoll, Christian Theobalt, Naveed Ahmed, Hans-Peter
  Seidel, and Sebastian Thrun.
\newblock Performance capture from sparse multi-view video.
\newblock In {\em ACM SIGGRAPH 2008 papers}, pages 1--10. 2008.

\bibitem{dong2014learning}
Chao Dong, Chen~Change Loy, Kaiming He, and Xiaoou Tang.
\newblock Learning a deep convolutional network for image super-resolution.
\newblock In {\em European conference on computer vision}, pages 184--199.
  Springer, 2014.

\bibitem{ferradans2014regularized}
Sira Ferradans, Nicolas Papadakis, Gabriel Peyr{\'e}, and Jean-Fran{\c{c}}ois
  Aujol.
\newblock Regularized discrete optimal transport.
\newblock {\em SIAM Journal on Imaging Sciences}, 7(3):1853--1882, 2014.

\bibitem{finlayson2015color}
Graham~D Finlayson, Michal Mackiewicz, and Anya Hurlbert.
\newblock Color correction using root-polynomial regression.
\newblock {\em IEEE Transactions on Image Processing}, 24(5):1460--1470, 2015.

\bibitem{freedman2010object}
Daniel Freedman and Pavel Kisilev.
\newblock Object-to-object color transfer: Optimal flows and smsp
  transformations.
\newblock In {\em 2010 IEEE Computer Society Conference on Computer Vision and
  Pattern Recognition}, pages 287--294. IEEE, 2010.

\bibitem{grogan2019l2}
Mairead Grogan and Rozenn Dahyot.
\newblock L2 divergence for robust colour transfer.
\newblock {\em Computer Vision and Image Understanding}, 181:39--49, 2019.

\bibitem{guo2019relightables}
Kaiwen Guo, Peter Lincoln, Philip Davidson, Jay Busch, Xueming Yu, Matt Whalen,
  Geoff Harvey, Sergio Orts-Escolano, Rohit Pandey, Jason Dourgarian, et~al.
\newblock The relightables: Volumetric performance capture of humans with
  realistic relighting.
\newblock {\em ACM Transactions on Graphics (TOG)}, 38(6):1--19, 2019.

\bibitem{haris2018deep}
Muhammad Haris, Gregory Shakhnarovich, and Norimichi Ukita.
\newblock Deep back-projection networks for super-resolution.
\newblock In {\em Proceedings of the IEEE conference on computer vision and
  pattern recognition}, pages 1664--1673, 2018.

\bibitem{he2018deep}
Mingming He, Dongdong Chen, Jing Liao, Pedro~V Sander, and Lu Yuan.
\newblock Deep exemplar-based colorization.
\newblock {\em ACM Transactions on Graphics (TOG)}, 37(4):1--16, 2018.

\bibitem{hu2019instance}
Guanqing Hu and James Clark.
\newblock Instance segmentation based semantic matting for compositing
  applications.
\newblock In {\em 2019 16th Conference on Computer and Robot Vision (CRV)},
  pages 135--142. IEEE, 2019.

\bibitem{huang2018deep}
Zeng Huang, Tianye Li, Weikai Chen, Yajie Zhao, Jun Xing, Chloe LeGendre,
  Linjie Luo, Chongyang Ma, and Hao Li.
\newblock Deep volumetric video from very sparse multi-view performance
  capture.
\newblock In {\em Proceedings of the European Conference on Computer Vision
  (ECCV)}, pages 336--354, 2018.

\bibitem{hwang2014color}
Youngbae Hwang, Joon-Young Lee, In So~Kweon, and Seon Joo~Kim.
\newblock Color transfer using probabilistic moving least squares.
\newblock In {\em Proceedings of the IEEE conference on computer vision and
  pattern recognition}, pages 3342--3349, 2014.

\bibitem{isola2017image}
Phillip Isola, Jun-Yan Zhu, Tinghui Zhou, and Alexei~A Efros.
\newblock Image-to-image translation with conditional adversarial networks.
\newblock In {\em Proceedings of the IEEE conference on computer vision and
  pattern recognition}, pages 1125--1134, 2017.

\bibitem{jeong2008object}
Kideog Jeong and Christopher Jaynes.
\newblock Object matching in disjoint cameras using a color transfer approach.
\newblock {\em Machine Vision and Applications}, 19(5-6):443--455, 2008.

\bibitem{joo2015panoptic}
Hanbyul Joo, Hao Liu, Lei Tan, Lin Gui, Bart Nabbe, Iain Matthews, Takeo
  Kanade, Shohei Nobuhara, and Yaser Sheikh.
\newblock Panoptic studio: A massively multiview system for social motion
  capture.
\newblock In {\em Proceedings of the IEEE International Conference on Computer
  Vision}, pages 3334--3342, 2015.

\bibitem{klinger1967vandermonde}
Allen Klinger.
\newblock The vandermonde matrix.
\newblock {\em The American Mathematical Monthly}, 74(5):571--574, 1967.

\bibitem{lahner2018deepwrinkles}
Zorah Lahner, Daniel Cremers, and Tony Tung.
\newblock Deepwrinkles: Accurate and realistic clothing modeling.
\newblock In {\em Proceedings of the European Conference on Computer Vision
  (ECCV)}, pages 667--684, 2018.

\bibitem{lanaras2018super}
Charis Lanaras, Jos{\'e} Bioucas-Dias, Silvano Galliani, Emmanuel Baltsavias,
  and Konrad Schindler.
\newblock Super-resolution of sentinel-2 images: Learning a globally applicable
  deep neural network.
\newblock {\em ISPRS Journal of Photogrammetry and Remote Sensing},
  146:305--319, 2018.

\bibitem{ledig2017photo}
Christian Ledig, Lucas Theis, Ferenc Husz{\'a}r, Jose Caballero, Andrew
  Cunningham, Alejandro Acosta, Andrew Aitken, Alykhan Tejani, Johannes Totz,
  Zehan Wang, et~al.
\newblock Photo-realistic single image super-resolution using a generative
  adversarial network.
\newblock In {\em Proceedings of the IEEE conference on computer vision and
  pattern recognition}, pages 4681--4690, 2017.

\bibitem{lee2019perceptual}
Wei-Yu Lee, Po-Yu Chuang, and Yu-Chiang~Frank Wang.
\newblock Perceptual quality preserving image super-resolution via channel
  attention.
\newblock In {\em ICASSP 2019-2019 IEEE International Conference on Acoustics,
  Speech and Signal Processing (ICASSP)}, pages 1737--1741. IEEE, 2019.

\bibitem{li2020survey}
Kai Li, Shenghao Yang, Runting Dong, Xiaoying Wang, and Jianqiang Huang.
\newblock Survey of single image super-resolution reconstruction.
\newblock {\em IET Image Processing}, 14(11):2273--2290, 2020.

\bibitem{li20193d}
Yawei Li, Vagia Tsiminaki, Radu Timofte, Marc Pollefeys, and Luc~Van Gool.
\newblock 3d appearance super-resolution with deep learning.
\newblock In {\em Proceedings of the IEEE Conference on Computer Vision and
  Pattern Recognition}, pages 9671--9680, 2019.

\bibitem{li2019feedback}
Zhen Li, Jinglei Yang, Zheng Liu, Xiaomin Yang, Gwanggil Jeon, and Wei Wu.
\newblock Feedback network for image super-resolution.
\newblock In {\em Proceedings of the IEEE Conference on Computer Vision and
  Pattern Recognition}, pages 3867--3876, 2019.

\bibitem{lim2017enhanced}
Bee Lim, Sanghyun Son, Heewon Kim, Seungjun Nah, and Kyoung Mu~Lee.
\newblock Enhanced deep residual networks for single image super-resolution.
\newblock In {\em Proceedings of the IEEE conference on computer vision and
  pattern recognition workshops}, pages 136--144, 2017.

\bibitem{Mei_2020_CVPR}
Yiqun Mei, Yuchen Fan, Yuqian Zhou, Lichao Huang, Thomas~S. Huang, and Honghui
  Shi.
\newblock Image super-resolution with cross-scale non-local attention and
  exhaustive self-exemplars mining.
\newblock In {\em Proceedings of the IEEE/CVF Conference on Computer Vision and
  Pattern Recognition (CVPR)}, June 2020.

\bibitem{oliveira2014probabilistic}
Miguel Oliveira, Angel~Domingo Sappa, and Vitor Santos.
\newblock A probabilistic approach for color correction in image mosaicking
  applications.
\newblock {\em IEEE Transactions on image Processing}, 24(2):508--523, 2014.

\bibitem{orts2016holoportation}
Sergio Orts-Escolano, Christoph Rhemann, Sean Fanello, Wayne Chang, Adarsh
  Kowdle, Yury Degtyarev, David Kim, Philip~L Davidson, Sameh Khamis, Mingsong
  Dou, et~al.
\newblock Holoportation: Virtual 3d teleportation in real-time.
\newblock In {\em Proceedings of the 29th Annual Symposium on User Interface
  Software and Technology}, pages 741--754, 2016.

\bibitem{park2016efficient}
Jaesik Park, Yu-Wing Tai, Sudipta~N Sinha, and In So~Kweon.
\newblock Efficient and robust color consistency for community photo
  collections.
\newblock In {\em Proceedings of the IEEE Conference on Computer Vision and
  Pattern Recognition}, pages 430--438, 2016.

\bibitem{pitie2007linear}
Fran{\c{c}}ois Piti{\'e} and Anil Kokaram.
\newblock The linear monge-kantorovitch linear colour mapping for example-based
  colour transfer.
\newblock 2007.

\bibitem{pitie2007automated}
Fran{\c{c}}ois Piti{\'e}, Anil~C Kokaram, and Rozenn Dahyot.
\newblock Automated colour grading using colour distribution transfer.
\newblock {\em Computer Vision and Image Understanding}, 107(1-2):123--137,
  2007.

\bibitem{pouli2011progressive}
Tania Pouli and Erik Reinhard.
\newblock Progressive color transfer for images of arbitrary dynamic range.
\newblock {\em Computers \& Graphics}, 35(1):67--80, 2011.

\bibitem{reinhard2001color}
Erik Reinhard, Michael Adhikhmin, Bruce Gooch, and Peter Shirley.
\newblock Color transfer between images.
\newblock {\em IEEE Computer graphics and applications}, 21(5):34--41, 2001.

\bibitem{richard2019learned}
Audrey Richard, Ian Cherabier, Martin~R Oswald, Vagia Tsiminaki, Marc
  Pollefeys, and Konrad Schindler.
\newblock Learned multi-view texture super-resolution.
\newblock In {\em 2019 International Conference on 3D Vision (3DV)}, pages
  533--543. IEEE, 2019.

\bibitem{starck2007surface}
Jonathan Starck and Adrian Hilton.
\newblock Surface capture for performance-based animation.
\newblock {\em IEEE computer graphics and applications}, 27(3):21--31, 2007.

\bibitem{Vasu_2018_ECCV_Workshops}
Subeesh Vasu, Nimisha Thekke~Madam, and A.~N. Rajagopalan.
\newblock Analyzing perception-distortion tradeoff using enhanced perceptual
  super-resolution network.
\newblock In {\em Proceedings of the European Conference on Computer Vision
  (ECCV) Workshops}, September 2018.

\bibitem{vlasic2009dynamic}
Daniel Vlasic, Pieter Peers, Ilya Baran, Paul Debevec, Jovan Popovi{\'c},
  Szymon Rusinkiewicz, and Wojciech Matusik.
\newblock Dynamic shape capture using multi-view photometric stereo.
\newblock In {\em ACM SIGGRAPH Asia 2009 papers}, pages 1--11. 2009.

\bibitem{wang2018esrgan}
Xintao Wang, Ke Yu, Shixiang Wu, Jinjin Gu, Yihao Liu, Chao Dong, Yu Qiao, and
  Chen Change~Loy.
\newblock Esrgan: Enhanced super-resolution generative adversarial networks.
\newblock In {\em Proceedings of the European Conference on Computer Vision
  (ECCV)}, pages 0--0, 2018.

\bibitem{xian2018texturegan}
Wenqi Xian, Patsorn Sangkloy, Varun Agrawal, Amit Raj, Jingwan Lu, Chen Fang,
  Fisher Yu, and James Hays.
\newblock Texturegan: Controlling deep image synthesis with texture patches.
\newblock In {\em Proceedings of the IEEE Conference on Computer Vision and
  Pattern Recognition}, pages 8456--8465, 2018.

\bibitem{xiang2009selective}
Yao Xiang, Beiji Zou, and Hong Li.
\newblock Selective color transfer with multi-source images.
\newblock {\em Pattern Recognition Letters}, 30(7):682--689, 2009.

\bibitem{Xie_2019_ICCV}
Haozhe Xie, Hongxun Yao, Xiaoshuai Sun, Shangchen Zhou, and Shengping Zhang.
\newblock Pix2vox: Context-aware 3d reconstruction from single and multi-view
  images.
\newblock In {\em Proceedings of the IEEE/CVF International Conference on
  Computer Vision (ICCV)}, October 2019.

\bibitem{yang2014single}
Chih-Yuan Yang, Chao Ma, and Ming-Hsuan Yang.
\newblock Single-image super-resolution: A benchmark.
\newblock In {\em European Conference on Computer Vision}, pages 372--386.
  Springer, 2014.

\bibitem{yang2020learning}
Fuzhi Yang, Huan Yang, Jianlong Fu, Hongtao Lu, and Baining Guo.
\newblock Learning texture transformer network for image super-resolution.
\newblock In {\em Proceedings of the IEEE/CVF Conference on Computer Vision and
  Pattern Recognition}, pages 5791--5800, 2020.

\bibitem{yang2015chi}
Wei Yang, Luhui Xu, Xiaopan Chen, Fengbin Zheng, and Yang Liu.
\newblock Chi-squared distance metric learning for histogram data.
\newblock {\em Mathematical Problems in Engineering}, 2015, 2015.

\bibitem{yuhas1992discrimination}
Roberta~H Yuhas, Alexander~FH Goetz, and Joe~W Boardman.
\newblock Discrimination among semi-arid landscape endmembers using the
  spectral angle mapper (sam) algorithm.
\newblock In {\em Proc. Summaries 3rd Annu. JPL Airborne Geosci. Workshop},
  volume~1, pages 147--149, 1992.

\bibitem{zhang2011fsim}
Lin Zhang, Lei Zhang, Xuanqin Mou, and David Zhang.
\newblock Fsim: A feature similarity index for image quality assessment.
\newblock {\em IEEE transactions on Image Processing}, 20(8):2378--2386, 2011.

\bibitem{zhang2014comparison}
Qianwen Zhang and Roxanne~L Canosa.
\newblock A comparison of histogram distance metrics for content-based image
  retrieval.
\newblock In {\em Imaging and Multimedia Analytics in a Web and Mobile World
  2014}, volume 9027, page 90270O. International Society for Optics and
  Photonics, 2014.

\bibitem{zhang2018image}
Yulun Zhang, Kunpeng Li, Kai Li, Lichen Wang, Bineng Zhong, and Yun Fu.
\newblock Image super-resolution using very deep residual channel attention
  networks.
\newblock In {\em Proceedings of the European Conference on Computer Vision
  (ECCV)}, pages 286--301, 2018.

\bibitem{zhang2019image}
Zhifei Zhang, Zhaowen Wang, Zhe Lin, and Hairong Qi.
\newblock Image super-resolution by neural texture transfer.
\newblock In {\em Proceedings of the IEEE Conference on Computer Vision and
  Pattern Recognition}, pages 7982--7991, 2019.

\bibitem{zheng2018crossnet}
Haitian Zheng, Mengqi Ji, Haoqian Wang, Yebin Liu, and Lu Fang.
\newblock Crossnet: An end-to-end reference-based super resolution network
  using cross-scale warping.
\newblock In {\em Proceedings of the European Conference on Computer Vision
  (ECCV)}, pages 88--104, 2018.

\bibitem{1284395}
{Zhou Wang}, A.~C. {Bovik}, H.~R. {Sheikh}, and E.~P. {Simoncelli}.
\newblock Image quality assessment: from error visibility to structural
  similarity.
\newblock {\em IEEE Transactions on Image Processing}, 13(4):600--612, 2004.

\bibitem{zhu2017unpaired}
Jun-Yan Zhu, Taesung Park, Phillip Isola, and Alexei~A Efros.
\newblock Unpaired image-to-image translation using cycle-consistent
  adversarial networks.
\newblock In {\em Proceedings of the IEEE international conference on computer
  vision}, pages 2223--2232, 2017.

\end{thebibliography}
}

\end{document}



\title{Super-Resolution Appearance Transfer for 4D Human Performances - Supplementary Material}

\author{Marco Pesavento \hspace{3cm} Marco Volino \hspace{3cm} Adrian Hilton \\ 
Centre for Vision, Speech and Signal Processing\\
University of Surrey, UK\\
{\tt\small \{m.pesavento,m.volino,a.hilton\}@surrey.ac.uk}
}
\maketitle
\thispagestyle{empty}

This document presents additional details and qualitative results for the SRAT pipeline introduced in the main paper. Section~\ref{sec:supcol} further explains Equation 2 of the paper. Section~\ref{sec:suptrain} explains the training procedure for RCAN~\cite{zhang2018image}.
Section~\ref{sec:evalmetric} defines the metrics used for evaluating our pipeline stages. Section~\ref{sec:limitation} presents the limitations of SRAT and Section~\ref{sec:subeval} offers further evaluations of the SRAT stages. Visual results of the colour mapping stage are shown in Section~\ref{ssec:supcolcorr}. Other results for the super-resolution (SR) stage are presented in Section~\ref{ssec:supSR}. In Section~\ref{ssec:supperf}, there are additional considerations on the performances of different configurations of SRAT. 
\section{Colour transfer algorithm}
\label{sec:supcol}
\noindent
To learn the colour transfer function $\phi_{\theta}(x)$ that aims to transfer the appearance from a high-resolution (HR) static capture of a human subject to a dynamic video performance capture of the same subject, we introduce a parametric energy function (Equation 2 in the main paper): 
\begin{equation}
\label{eq:1}
\theta=\frac{1}{N}\sum_{l=1}^{N}\argmin_{\theta_l}\{||p_f||^2-2<p_f|p_I>\}
\end{equation}
This function is minimized with gradient descent algorithm where:
\begin{equation}
\label{eq:2}
    \scriptstyle
    ||p_f||^2=\sum_{i=1}^{K}\sum_{k=1}^{K} \mathcal{N}(0;\phi_\theta(\mu_f^{(k)})-\phi_\theta(\mu_f^{(i)}),2h^2I)\pi_f^{(k)}\pi_f^{(i)}
\end{equation}
\begin{equation}
\label{eq:3}
    \scriptstyle
   <p_f|p_I>=  \sum_{i=1}^{K}\sum_{k=1}^{K}\mathcal{N}(0;\phi_\theta(\mu_f^{(k)})-\mu_I^{(i)},2h^2I)\pi_f^{(k)}\pi_I^{(i)}
\end{equation}
$p_f$ is the distribution of the input frame image $I_{LR}$ while $p_I$ is the distribution of the correspondent HR image $I_{HR}$. These two images are paired during the first step of the second stage of SRAT (couples identification).
\\$p_f$ is a Gaussian mixture distribution of parameter $K$ defined by the Normal distribution $\mathcal{N}(0;\phi_\theta(\mu_f),h^2I)$ computed at 0 with parameterised mean $\phi_\theta(\mu_f)$ and covariance $h^2I$ ($h$ is a control bandwidth and $I$ is the identity matrix).
$p_I$ is a Gaussian Mixture distribution of parameter $K$ as well, but its mean $\mu_I$ is not parameterised.  
\\In Equation~\ref{eq:2}, the norm is computed as the subtraction of the parameterised mean of two cluster $i$ and $k$ of $p_f$ while in Equation~\ref{eq:3} the mean of $p_I$ is subtracted to the parameterised one of $p_f$. 
The weights ${\pi_f^{(i)}}_{i=1,...,K}$ as well as ${\pi_I^{(k)}}_{k=1,...,K}$ are chosen equiprobable with ${\pi_f^{(i)}}=1/K$ and ${\pi_I^{(k)}}=1/K$ respectively. 
\\$K$ is a fixed number defining the K variable of the K-means algorithm. K-means selects $K$ clusters in the colours of the input images to define their Gaussian mixture distributions $p_f$ and $p_I$ before the minimization step~\cite{grogan2019l2}.
\\In other words, the algorithm first selects $K$ clusters in the colours of the target ($I_{LR}$) and of the reference ($I_{HR}$) image. Then, it models the colour distribution as a Gaussian mixture distribution whose mean is parameterised only for $I_{LR}$. Finally, the energy function of Equation~\ref{eq:1} is minimized and $\phi_{\theta}(x)$ is defined from a set of input couples of $I_{LR}$ and $I_{HR}$.
\section{Implementation of RCAN}
\label{sec:suptrain}
\noindent
In this work we use RCAN-style network~\cite{zhang2018image} to super resolve texture map. In order to achieve this, we adapt the training protocol and utilise datasets of human texture maps to train and fine-tune RCAN.
\\\indent\textbf{Training datasets: }RCAN presents a very deep architecture and the larger the architecture, the more data is needed to produce viable results. A common problem derives from the lack of human texture maps available for training. 
\\We first train RCAN with a dataset made of texture maps of 7 human models downloaded from~\cite{renderpeople}. To augment the training data and to ease the training, we crop the texture maps into patches with the size of 256x256. The final dataset presents 1872 patches. 
\\To enhance RCAN performances, we fine-tune it with original texture maps of the input model retrieved in a pre-processing stage. In particular, we use 410 texture maps for \textit{SingleF} model and 441 for \textit{SingleM}.
For every different input model of SRAT, RCAN can be fine-tuned with the original texture maps of the specific model.
\\\indent\textbf{Training process: }RCAN was trained for 800 epochs with the patches of texture maps and then fine-tuned for 200 epochs with the original texture maps of the input model. 
\\In each training batch, 16 low-resolution (LR) color patches (48x48 in size) are extracted as inputs. The learning rate is initially set to $10^{-4}$ and then halved every $10^5$ iterations of back-propagation. ADAM optimizer~\cite{kingma2014adam} is used with $\beta_1$$ =$$ 0.9$, $\beta_2$$ =$$ 0.999$ and $\epsilon=10^8$.
\section{Evaluation metrics}
\label{sec:evalmetric}
\indent\textbf{Colour transfer stage: }to evaluate the effect of the colour transfer from the HR images to the input frames, we consider their normalized colour histograms. We compute the Jensen-Shannon (JS) divergence and the Chi-Squared ($\chi^2$) distance. Specifically, we measure the dissimilarity of the colour histogram of a frame with the colour histogram of the HR image which was paired with the frame in the colour correction step of the colour mapping stage of SRAT. 
The JS divergence measures the similarity of two statistical distributions by subtracting the average of the two distribution entropies with the entropy of their average~\cite{zhang2014comparison}:
\begin{equation}
    \textstyle
    JS(H,H')=\textstyle \sum_{l=1}^{N}(H_i\log(\frac{2H_i}{H_i+H_i'}+H_i'\log(\frac{2H_i'}{H_i+H_i'})))
\end{equation}
The $\chi^2$ distance weights with higher importance the dissimilarity between two small bins of the histograms~\cite{zhang2014comparison}:
\begin{equation}
    \chi^2(H,H')=\sum_{l=1}^{N}\frac{2(H_i-H_i')^2}{H_i+H_i'}
\end{equation}
where, for both the equations, $H$ is the normalized colour histogram of the frame, $H'$ is the normalized colour histogram of the HR image, $N$ is the the number of bins of each histogram and $H_i$ is the value of the $i^{th}$ bin of $H$~\cite{zhang2014comparison}.
\\We select $N=64$. The presented results are computed considering all the frames of \textit{SingleF} (7040) and \textit{SingleM} (7520) models.
\\\indent\textbf{Texture-map SR stage: }the SR texure maps are evaluated with PSNR and SSIM~\cite{1284395} on Y channel of transformed YCbCr space. These metrics are computed by testing the network with 104 texture maps of \textit{SingleF} and with 113 texture maps of \textit{SingleM}.
\section{Limitations} 
\label{sec:limitation}
\noindent
SRAT only enhances the appearance of the model. Therefore, if the initial reconstruction produces low quality 3D shapes, our pipeline is not able to modify the geometry of the model and the final output will still have a low quality shape. Another limitation is during the couple identification step of the colour mapping stage when Densepose is not able to detect the human model. The partial texture map is in this case a black image and the couple cannot be created. If the selected couples for learning the colour transfer function do not cover the surface entirely, the corrected frames present unnatural colours. Figure~\ref{fig:limitation} illustrates some of these limitations.
\begin{table}[!]
    \centering
    \resizebox{1\linewidth}{!}{
\begin{tabular}{ccc}
             \includegraphics[width=1.26in]{img/fail1.png} & \includegraphics[width=1.1in]{img/fail2_new.png} & \includegraphics[width=1.1in]{img/fail3_new.png}\\
             \scriptsize{(a)} & \scriptsize{(b)} & \scriptsize{(c)} \\
            
        \end{tabular}}
    \vspace{-1.0em}
     \captionof{figure}{Failure cases: (a) 4D reconstruction, (b) \textit{SingleM} colour transfer, (c) \textit{SingleF} colour transfer.}
     \vspace{-1.5em}
\label{fig:limitation}
\end{table}
\section{Further results evaluation}
\label{sec:subeval}
\noindent
In this section, additional visual results are shown for the evaluation studies of the main paper. A detailed analysis of the performances of different configurations is undertaken as well.
\subsection{Colour mapping stage evaluation}
\label{ssec:supcolcorr}
\indent\textbf{Input couple selection: }
visual results of failure cases caused by not giving as input to the colour transfer algorithm enough couples to cover the whole surface are shown in Figure~\ref{fig:inpufail1} for 1 input couple and Figure~\ref{fig:inpufail2} for 2 input couples for \textit{SingleF}. For \textit{SingleM} model, the results with 1 couple as input are illustrated in Figure~\ref{fig:inpufail3} and with 2 couples in Figure~\ref{fig:inpufail4}. In the learning stage not all the colours of the model are present in the input couples and in the reference palette some colours of the model are therefore missing. The outputs show how the colours of the specific parts of the subject that were not seen during the learning stage are unnatural and influenced by the colours of other parts of the subject.
\\\indent\textbf{Comparison with TPS~\cite{grogan2019l2}: }our colour transfer algorithm is based on TPS~\cite{grogan2019l2}. This method, as explained in the main paper, learns the colour transfer function from only a pair of target and reference image. Therefore, the algorithm learns a different function for every input frame because it models the colour distributions applying K-means algorithm. The learned colour transfer function can vary every time K-means is applied.
The colours of the corrected outputs may differ for consecutive frames as well as for the same frame of different cameras. Examples of the former case are shown in Figure~\ref{fig:divfail1} for \textit{SingleF} and Figure~\ref{fig:divfail3} for \textit{SingleM}. The second failure case is illustrated in Figure~\ref{fig:divfail2} for \textit{SingleF} and Figure~\ref{fig:divfail4} for \textit{SingleM}. If TPS is applied, the colours of the clothes and face of the models change in consecutive frames and when they are acquired with different cameras. On the contrary, the colours do not change if our method is applied in the aforementioned cases . 
\subsection{Texture-map super resolution stage evaluation: }
\label{ssec:supSR}
\\\indent\textbf{Different training models of RCAN: }Figure~\ref{fig:supSR1} shows the outputs of all the considered training models of RCAN for \textit{SingleF} while Figure~\ref{fig:supSR12} for \textit{SingleM}. A portion of the texture-map and its correspondent heatmap are illustrated. The heatmap highlights the dissimilarity between the ground-truth and the SR texture map: in a scale from blue to red colours, the blue one indicates that two correspondent pixels of the ground-truth and the SR texture maps are the most similar while the red colour represents the pixels which are most dissimilar. The heatmap of the best case (1b) adopted in our pipeline presents more blue and less green/red pixels compared to the others.  
\\\indent\textbf{Comparison with SR networks: }Figure~\ref{fig:supSR2} shows a portion of the texture-map and its correspondent heatmap for \textit{SingleF} and Figure~\ref{fig:supSR22} for \textit{SingleM} for the considered SR networks. Also in this case, the SRAT heatmap shows more blue pixels for both the models.
\\\indent\textbf{Final model outputs: }Figure~\ref{fig:supout} shows details of the \textit{SingleF} 3D model when the output SR texture maps of SRAT are rendered to the meshes. Figure~\ref{fig:supout2} illustrates \textit{SingleM}. As it is possible to see in the hairs and in the band of \textit{SingleF} and in the button of \textit{SingleM}, the details of the super resolved models appear less blurry than their LR version.
\\Figure~\ref{fig:supout3} shows the output 3D models of SRAT, whose appearance is significantly enhanced compared to their original version, with brighter colours.
\begin{table}[]
\resizebox{1.0\linewidth}{!}{
\begin{tabular}{|cccccc|}
\hline
\multicolumn{1}{|c|}{\textbf{Config.}} & \multicolumn{1}{c|}{\begin{tabular}[c]{@{}c@{}}\textbf{Colour}\\  \textbf{transfer} \\(s)\end{tabular}} & \multicolumn{1}{c|}{\begin{tabular}[c]{@{}c@{}}\textbf{Super}\\  \textbf{resolution} \\(s) \end{tabular}} & \multicolumn{1}{c|}{\begin{tabular}[c]{@{}c@{}}\textbf{Average time}\\ \textbf{per frame}\\(s)\end{tabular}} & \multicolumn{1}{c|}{\begin{tabular}[c]{@{}c@{}}\textbf{Total  time}\\ \textbf{SingleF} \\ (hours)\end{tabular} 
} & \multicolumn{1}{c|}{\begin{tabular}[c]{@{}c@{}}\textbf{Total time}\\ \textbf{SingleM} \\ (hours)\end{tabular} } \\ \hline
SRAT (ours)    & 14.13                                                                        & 128.49 & 354.57                                                                      & 43.33                                                                   & 46.29                                                                  \\
1              & 8.45                                                                         & 128.49 & 136.94                                                                      & 16.73                                                                       & 17.87                                                                   \\
2              & 35.16                                                                        & 128.49 & 63.65                                                                      & 20                                                                      & 21.36                                                                    \\
3              & 14.13                                                                        & 197.19  & 3381.12                                                                 & 413.24                                                                & 441.42                                                               \\
4              & 56.71                                                                        & 197.19 & 4062.4                                                                  & 496.51                                                                 & 530.36                                                                 \\
5              & 35.16                                                                        & 197.19 & 759.75                                                                     & 389.91                                                                  & 416.49        \\                      \hline                                   
\end{tabular}}
\caption{Performance of SRAT pipeline configurations.}
\label{tbl:config}
\end{table}
\subsection{Different configurations of SRAT}
\label{ssec:supperf}
\noindent
Table~\ref{tbl:config} presents the performances of the studied configurations of SRAT (Section 4.4 of the main paper). We measure the time in seconds (s) to complete the colour transfer step and the SR texture map stage for both \textit{SingleF} and \textit{SingleM} models. The average time to process a single frame on the two stages is also presented. In this case, there are 16 input images (one for each camera) and 1 texture map per frame. 
\\In addition, we compute the total time (in hours) to process all the frames for each performance. \textit{SingleF} has 440 frames per camera and 7040 total frame images and 440 texture maps in total. \textit{SingleM} presents 470 frames per camera (7520 frame images and 470 texture maps).
The size of the input frame images is 3840x2160 pixels (7680x4320 pixels when super resolved by a factor of 2) and the texture map size is 2048x2048 pixels (4096x4096 pixels when super-resolved by a factor of 2). For the super-resolution stage an NVIDIA GeForce RTX 2070 was used. If the colour mapping stage was processed with multiple CPUs and the SR stage with multiple GPUs, the processing time would be lower than the one shown in Table~\ref{tbl:config}. We do not measure the time of the other stages because is constant in all the configurations. The fastest configurations are the 1st and the 2nd but their outputs present visible artefacts as shown in the main paper. The proposed SRAT pipeline is the third fastest and produces the best visual results.
 {\small
 \bibliographystyle{ieee}
 \bibliography{egbib}
 }